\documentclass{article}
\usepackage{ismir,amsmath,cite,url}
\usepackage{graphicx}
\usepackage{color}
\usepackage{multirow}
\usepackage{float}

\title{EXPLOITING SYNCHRONIZED LYRICS AND VOCAL FEATURES FOR MUSIC EMOTION DETECTION}

\multauthor
{Loreto Parisi$^1$ \hspace{1cm} Simone Francia$^1$ \hspace{1cm} {Silvio Olivastri$^2$ \hspace{1cm} Maria Stella Tavella$^1$}} 
{$^1$ Musixmatch \hspace{1cm} $^2$ AI Labs \\
{\tt\small loreto@musixmatch.com, simone.francia@musixmatch.com} \\
{\tt\small silvio@ailabs.it, stella@musixmatch.com}}
\def\authorname{Loreto Parisi, Simone Francia, Silvio Olivastri, Maria Stella Tavella}

\begin{document}

\maketitle

\begin{abstract}
\textbf{One of the key points in music recommendation is authoring engaging playlists according to sentiment and emotions. While previous works were mostly based on audio for music discovery and playlists generation, we take advantage of our synchronized lyrics dataset to combine text representations and music features in a novel way; we therefore introduce the Synchronized Lyrics Emotion Dataset. Unlike other approaches that randomly exploited the audio samples and the whole text, our data is split according to the temporal information provided by the synchronization between lyrics and audio. 
This work shows a comparison between text-based and audio-based deep learning classification models using different techniques from Natural Language Processing and Music Information Retrieval domains.
From the experiments on audio we conclude that using vocals only, instead of the whole audio data improves the overall performances of the audio classifier. In the lyrics experiments we exploit the state-of-the-art word representations applied to the main Deep Learning architectures available in literature.
In our benchmarks the results show how the Bilinear LSTM classifier with Attention based on fastText word embedding performs better than the CNN applied on audio.}
\end{abstract}
\section{Introduction}\label{sec:introduction}

Music Emotion Recognition (MER) refers to the task of finding a relationship between music and human emotions \cite{Yang:2011, Kim:2010}. Nowadays, this type of analysis is becoming more and more popular, music streaming providers are finding very helpful to present users with musical collections organized according to their feelings. 
The problem of Music Emotion Recognition was proposed for the first time in the Music Information Retrieval (MIR) community in 2007, during the annual Music Information Research Evaluation eXchange (MIREX) \cite{Hu:2008}. 
Audio and lyrics represent the two main sources from which it is possible to obtain low and high-level features that can accurately describe human moods and emotions perceived while listening to music.
An equivalent task can be performed in the area of Natural Language Processing (NLP) analyzing the text information of a song by labeling a sentence, in our case one or more lyrics lines, with the emotion associated to what it expresses.
A typical MER approach consists in training a classifier using various representations of the acoustical properties of a musical excerpt such as: timbre, rhythm and harmony \cite{Katayose:1998, Laar:2006}. 
Support Vector Machines are employed with good results also for multilabel classification \cite{Li:2003}, more recently also Convolutional Neural Networks were used in this field \cite{Yang:2012}. Lyrics-based approaches, on the other hand, make use of Recurrent Neural Networks architectures (like LSTM \cite{Hochreiter:1997}) for performing text classification \cite{Zhou:2016, Yang:2016}.
The idea of using lyrics combined with voice only audio signals is done in \cite{Lee:2018}, where emotion recognition is performed by using textual and speech data, instead of visual ones.
Measuring and assigning emotions to music is not a straightforward task: the sentiment/mood associated with a song can be derived by a combination of many features, moreover, emotions expressed by a musical excerpt and by its corresponding lyrics do not always match, also, the annotation of mood tags to a song turns out to be highly changing over the song duration \cite{Caetano:2013} and therefore one song can be associated with more than one emotion \cite{Yang:2012}.
There happens to be no unified emotion representation in the field of MER, meaning that there is no consensus upon which and how many labels are used and if emotions should be considered as categorical or continuous, moreover, emotion annotation has been carried on in different ways over the years, depending on the study and experiments conducted.
For this reason, researches have developed many different approaches and results visualization that is hard to track a precise state-of-the-art for this MIR task \cite{Yang:2012}.
In this work, our focus is on describing various approaches for performing emotion classification by analyzing lyrics and audio independently but in a synchronized fashion, as the lyric lines correspond to the portion of audio in which those same lines are sung and the audio is pre-processed using source separation techniques in order to separate the vocal part from the mixture of instruments and retaining the vocal tracks only.
\section{Related Works}\label{sec:related_works}

In this section we provide a description of musical emotion representation and the techniques for performing music emotion classification using audio and lyrics, illustrating various word embedding techniques for text classification. Finally we detail our proposed approach.

\subsection{Representing musical emotions}
\begin{figure}
\centering
\includegraphics[width=1\linewidth]{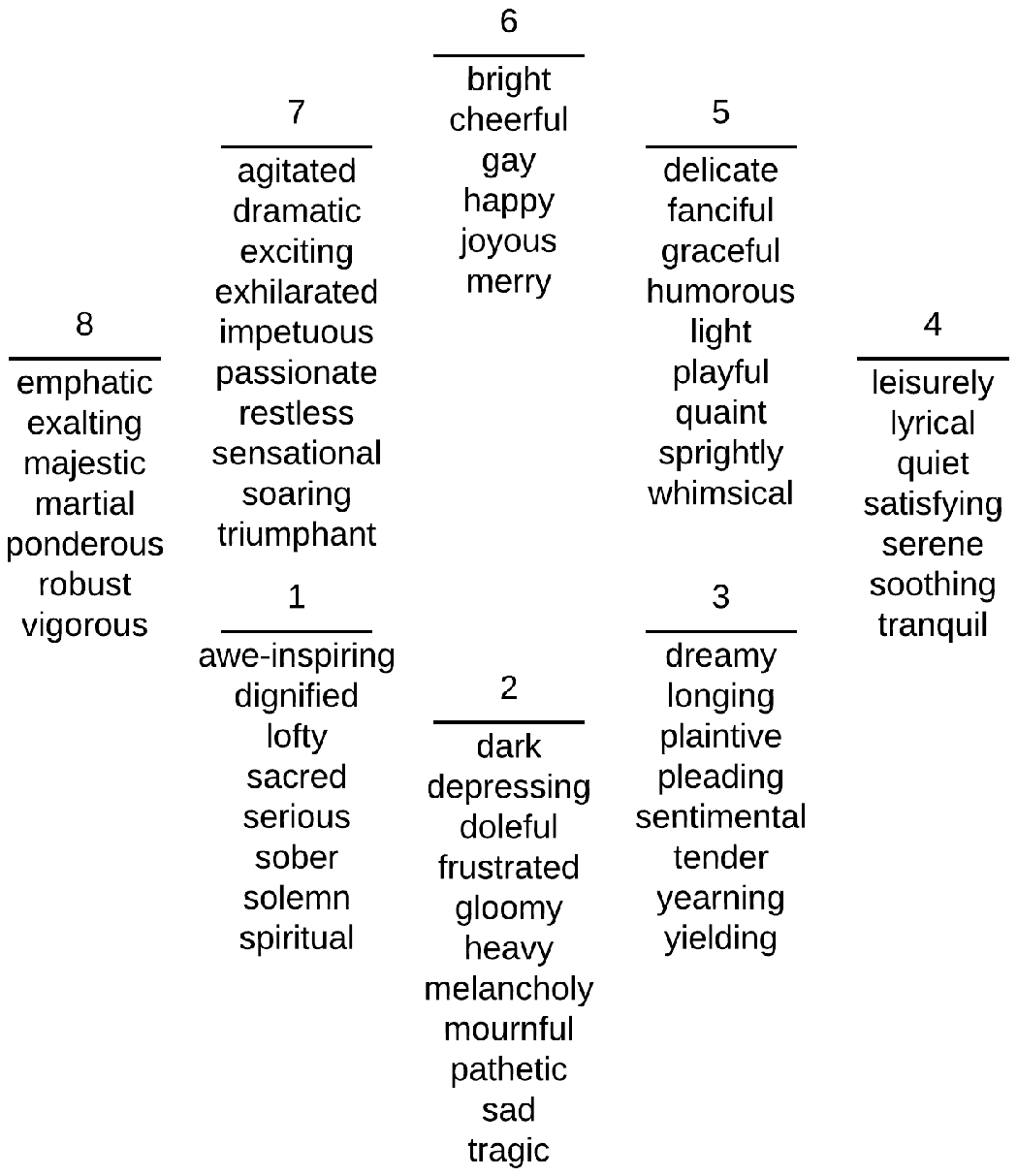}
\caption{Hevner Adjective Clusters. Sixtysix words arranged in eight clusters describing a variety of moods.}
\label{fig:hevner}
\end{figure}

\begin{figure}
\centering
  \includegraphics[width=1\linewidth]{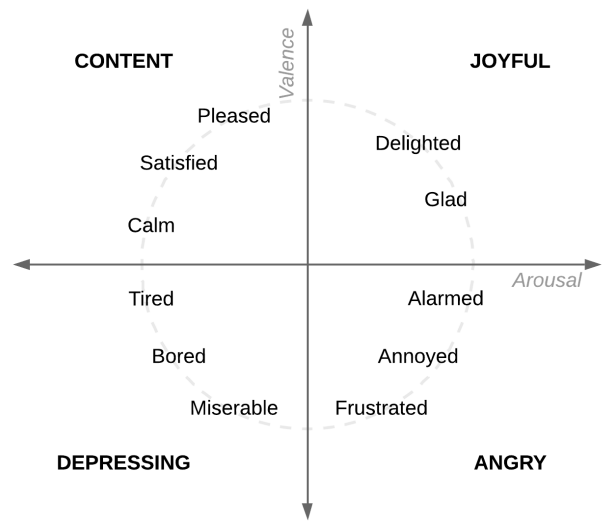}
  \caption{Russell's Valence-Arousal Space. A two-dimensional space representing mood states as continuous numerical values. The combination of valence-arousal values represent different emotions depending on their coordinates on the 2D space.}
  \label{fig:vaspace}
\end{figure}

Many studies were conducted for the representation of musical emotions also in the field of psychology. Despite cross-cultural studies suggesting that there may be universal psychophysical and emotional cues that transcend language and acculturation \cite{Kim:2010}, there exist several problems in recognizing moods from music. In fact, one of the main difficulties in recognizing musical mood is the ambiguity of human emotions. Different people perceive and feel emotions induced/expressed by music in many distinct ways. Also, their individual way of expressing them using adjectives is biased by a large number of variables and factors, such as the structure of music, the previous experience of the listener, their training, knowledge and psychological condition \cite{Hevner:1936}.
Music-IR systems use either categorical descriptions or parametric models of emotion for classification or recognition. Categorical approaches for the representation of emotions comprehend the finding and the organization of a set of adjectives/tags/labels that are emotional descriptors, based on their relevance and connection with music. One of the first studies concerning the aforementioned approach is the one conducted by Hevner and published in 1936, in which the candidates of the experiment were asked to choose from a list of 66 adjectives arranged in 8 groups \cite{Hevner:1936} as shown in Figure \ref{fig:hevner}.
Other research, such as the one conducted by Russell \cite{Russell:1980}, suggests that mood can be scaled and measured by a continuum of descriptors. In this scenario, sets of mood descriptors are organized into low-dimensional models, one of those models is the Valence-Arousal (V-A) space (Figure \ref{fig:vaspace}), in which emotions are organized on a plane along independent axes of Arousal (intensity, energy) and Valence (pleasantness), ranging from positive to negative.

\subsection{Music Emotion Classification}

MER can be approached either as a classification or regression problem in which a musical excerpt or a lyrics sentence is annotated with one or multiple emotion labels or with continuous values such as Valence-Arousal. It starts from the employment of certain acoustic features that resemble timbre, rhythm, harmony and other estimators in order to train Support Vector Machines at classifying the moods \cite{Li:2003}. Other techniques examine the use of Gaussian Mixture Models \cite{Lu:2006, Liu:2003} and Naive Bayes classifiers. The features involved for training those classifiers are computed based on the short time Fourier transform (STFT) for each frame of the sound. The most used features are the Mel-frequency cepstral coefficients (MFCC), spectral features such as the spectral centroid and the spectral flux are important for representing the timbral characteristic of the sound \cite{Tzanetakis:2002}.
Neural Networks are employed in \cite{Feng:2003}, where the tempo of the song is  computed by means of a multiple agents approach and, starting from other features computed on its statistics, a simple BP neural network classifier is trained to detect the mood.
It is not trivial to understand which features are more relevant for the task, therefore feature engineering has been recently carried on with the use of deep architectures (known in this context as feature learning) using either spectral representations of the audio (spectrograms or mel-spectrograms), in \cite{Choi:2016} feeded to a deep Fully Convolutional Networks (FCN) consisting in convolutional and subsampling layers without any fully-connected layer or directly using the raw audio signal as input to the network classifier \cite{Dieleman:2014}. 

Emotion recognition can also be addressed by using a lyrics-based approach. In \cite{Dakshina:2014} the probability of various emotions of a song are computed using Latent Dirichlet Allocation (LDA) based on latent semantic topics. In \cite{Malheiro:2016} the Russell's emotion model is employed and a keyword-based approach is used for classifying each sentence (verse).

Other works see the combination of audio-based classification with lyrics-based classification in order to improve the informative content of their vector representations and therefore the performances of the classification, as in \cite{Yang:2004, Schuller:2010, Bhattacharya:2018, Laurier:2008}. In \cite{Delbouys:2018} a 100-dimensional word2vec embedding trained on 1.6 million lyrics is tested comparing several architectures (GRU, LSTM, ConvNets). Not only the unified feature set results to be an advantage, but also the synchronized combination of both, as suggested in \cite{Delbouys:2018}. In \cite{Lee:2018} speech and text data are used in a fused manner for emotion recognition.

\subsection{Word Embedding and Text Classification}

In the field of NLP, text is usually converted into Bag of Words (BoW), Term Frequency–Inverse Document Frequency (TF-IDF) and, more recently, highly complex vector representations. In fact in the last few years, Word Embeddings have become an essential part of any Deep-Learning-based Natural Language Processing system representing the state-of-the-art for a large variety of classification task models.
Word Embeddings are pre-trained on a large corpus and can be fine-tuned to automatically improve their performance by incorporating some general representations. The Word2Vec method based on skip-gram \cite{Mikolov:2013}, had a large impact and enabled efficient training of dense word representations and a straightforward integration into downstream models. 
\cite{Kim:2016, Lample:2016, Yu:2017} added subword-level information and augmented word embeddings with character information for their relative applications. Later works \cite{Wieting:2016, Bojanowski:2017} showed that incorporating pre-trained embeddings character n-grams features provides more powerful results than composition functions over individual characters for several NLP tasks. Character n-grams are in particular efficient and also form the basis of Facebook's fastText classifier \cite{Joulin:2017_2, Joulin:2017, Bojanowski:2017}.
The most recent approaches \cite{Peters:2018, Devlin:2018} exploit contextual information extracted from bidirectional Deep Neural Models for which a different representation is assigned to each word and it is a function of the part of text to which the word belongs to, gaining state-of-the-art results for most NLP tasks. 
\cite{Howard:2018} achieves relevant results and is essentially a method to enable transfer learning for any NLP task without having to train models from scratch.
Regarding the prediction models for text classification, LSTM and RNN including all possible model Attention-based variants \cite{Zhou:2016} have represented for years the milestone for solving sequence learning, text classification \cite{Yang:2016} and machine translation \cite{Cho:2014}; other works show that CNN can be a good approach to solve NLP task too \cite{Kim:2014}.
In the last few years Transformer \cite{Vaswani:2017} outperformed both recurrent and convolutional approaches for language understanding, in particular on machine translation and language modeling. 

\subsection{Proposed Approach}

We built a new emotion dataset containing synchronized lyrics and audio. Our labels consist in 5 discrete crowd-based adjectives, inspired by the Hevner emotion representation retaining just the basics emotions as in \cite{Ekman:1999}. Our aim is to perform emotion classification using lyrics and audio independently but in a synchronized manner. We have analyzed the performances of different text embedding methods and used contextual feature extraction such as ELMo and  BERT combined with various classifiers. We have exploited novel WaveNet \cite{Lluis:2018, Stoller:2018} techniques for separating singing voice from the audio and used a Convolutional Neural Network for emotion classification.
\section{Synchronized Lyrics Emotion Dataset}\label{sec:dataset}

Previous methods combining text and audio for classification purposes were randomly analyzing 30 seconds long segments from the middle of the audio file \cite{Lu:2006}. Starting from the idea that the mood can change over time in an audio excerpt, as well as a song might express different moods depending on its lyrics and sentences, we exploit our time-synchronized data and build a novel dataset, the \textit{Synchronized Lyrics Emotion Dataset}, containing synchronized lyrics collected by the  Musixmatch\footnote{\texttt{https://developer.musixmatch.com/}} platform having start and duration times in which certain lyrics lines are sung.
Musical lyrics are the transcribed words of a song, therefore synchronization is a temporal information that establishes a direct connection between the text and its corresponding audio event interval, in other words, synchronizing lyrics is about storing information on the instance of time (in the audio) at which every lyric line starts and ends. 
Each sample of the dataset consists of 5 values: the track ID, the lyrics, start/end time information related to the audio file segments and its relative mood label. The 5 collected emotion labels are shown in Table \ref{table:moods} together with their distribution in the dataset.
Each audio segment is a slice of music where the corresponding text is sung and it is tagged with a mood label. Since we need a consistent dimension for our audio features, we chose the audio/text segments to be approximately 30 second long.
For simplicity, we distinguish three types of segments: \textit{intro, synch, outro}.
The intro is the portion of a song that goes from the beginning of the song until the starting of the first sung line. The synch part goes from the beginning of the first sung line until the last one. The outro is a segment that starts from the end of the last sung line until the end of the song. Usually, intro and outro do not contain vocals and, in order to fulfil a consistent analysis, we do not take into account those when analyzing the audio, since our goal is to exploit the synchronizations and therefore to analyze only portion of audio containing vocals.
In Table \ref{table:1}, we show some statistics on the aforementioned segments. 
As an example, in Table \ref{table:2} we show a sample row of the described dataset.

\begin{table}[ht]
\centering
\begin{tabular}{c|c}
\textbf{EMOTION} & \textbf{\%} \\ \hline
sadness & 43.9 \\  
joy & 40.2 \\ 
fear & 7.7 \\ 
anger & 6.0 \\ 
disgust & 2.1
\end{tabular}
\caption{List of Emotions and their distribution in the Dataset.}
\label{table:moods}
\end{table}

\begin{table}[hbt!]
\centering
\begin{tabular}{l|l|l|l|l}
 & \shortstack{\textbf{sync} \\ \textbf{intro outro}} 
 & \textbf{synch} 
& \textbf{intro}
& \textbf{outro} \\ \hline
min  & 0 & 0.25 & 0 & 0.83 \\
max  & 41.37 & 16.07  & 35.56 & 41.37 \\ 
mean  & 5.46 & 4.72 & 17.62 & 22.68 \\ 
std  & 8.89 & 6.03 & 17.04 & 35.81 \\ 
perc\_70 & 4.92 & 4.67 & 23.94 & 21.54 \\ 
perc\_90  & 9.03 & 7.97 & 23.94 & 21.54
\end{tabular}
\caption{Statistics on the Synchronized Lyrics Emotion Dataset. The first column collects statistics on the synch, intro and outro. The other columns show statistics for synch without intro and outro, only intro and only outro respectively. }
\label{table:1}
\end{table}

\begin{table}[ht]
\centering
\begin{tabular}{p{1.3cm}|p{6cm}}
  \textbf{id} & 124391 \\ \hline
  \textbf{text} & So stay right here with me I know you say its been a struggle we've lost our touch we have dreamt too much maybe that's the trouble \\ \hline
  \textbf{start} & 65.79 \\ \hline
  \textbf{end} & 99.14 \\ \hline
  \textbf{emotion} & sadness
 \end{tabular}
 \caption{Sample row from the Synchronized Lyrics Emotion Dataset. Each row contains 5 fields: the ID of the track, the lyrics line, start and end times and the emotion label.}
 \label{table:2}
\end{table}

\section{Models}\label{sec:models}
This section is divided in two parts: in the first, we present the methods we implemented for resolving text-based Lyrics Classification. In the second, we describe how we approached the problem by means of an audio-based Convolutional Neural Network.

\subsection{Text-based}
\begin{figure}
  \includegraphics[width=\linewidth]{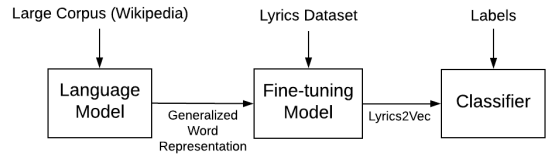}
  \caption{High Level Approach to model Lyrics General fine-tuning. From Wikipedia pre-trained language representation we obtain the Language Model fine-tuned on lyrics, Lyrics2Vec, with which a classifier is trained.}
  \label{fig:lyrics_pipeline}
\end{figure}

\begin{figure}
  \includegraphics[width=\linewidth]{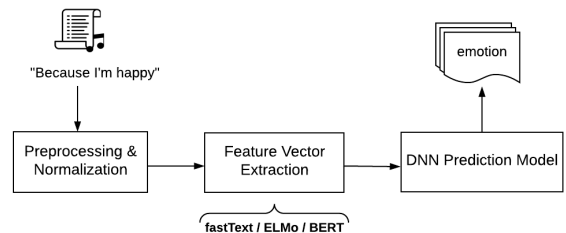}

  \caption{Lyrics Prediction Task Pipeline. Inputs of the pipeline are rows of the time-synchronized lyrics for which, after a text pre-processing and normalization phase, the embedding is calculated and it is used as input of a Deep Neural Network prediction model. }
  \label{fig:lyrics_method}
\end{figure}

We tested multiple word feature representations, our idea is to feed Deep Neural Network models with generic word embedding trained on a general corpus and, at the later stage, fine-tune all parameters in order to give the model the robustness of large dictionaries but at the same time specialized on our lyrics corpus, as Figure \ref{fig:lyrics_pipeline} shows. The main goal is to find, among all possible word representations, the one that fits better for lyrics emotion classification.
Feature Vector Representation that we chose to test are fastText \cite{Joulin:2017_2}, ELMo \cite{Peters:2018} and BERT \cite{Devlin:2018}: further, we combined different models for predicting lyrics emotions with all possible embeddings, as Figure \ref{fig:lyrics_method} shows. For every type of embedding we chose a way to extract an initial representation of every lyrics section.

\subsubsection{fastText}
In order to get a fixed representation of text in fastText, we set an embedding dimension of 300 for each word and then we average all the embeddings corresponding to every word and n-gram that appear in the input.

\begin{equation}\label{eq:fasttext_embedding}
E_{fastText}[l] = \frac{1}{M} \sum_{i}^{M} E_{fastText}[s_i]
\end{equation}

where $M$ is the number of sub-words ($s$) inside a lyrics ($l$) and $E_{fastText}$ is the application of embedding extraction from a general text. During training, several approaches were used: an initial fastText embedding with different Deep Neural Network in order to find which RNN cell architecture between LSTM and GRU is more suitable for solving text classification problems for lyrics. We also add a bidirectional version of both LSTM and GRU cell trying to understand if the analysis of text in two directions increases the performance of the classifier. In addition, layers for Attention-based \cite{Vaswani:2017} classification are added to our models: LSTM, BiLSTM \cite{Zhou:2016}, GRU, BiGRU.

\subsubsection{ELMo}
For ELMo we apply two different approaches for extracting the embeddings and consequently we use two different models for performing the classification. The first ELMo+LSTM applies, as initial embedding, the weighted sum of the hidden layers. This embedding has shape $[batch\_size, max\_length, 1024]$ and has been fine-tuned together with an LSTM followed by two dense layers for prediction task. In order to prevent overfitting, we provide a Batch Normalization layer before LSTM layer and we apply a $dropout rate$ 0.2 for LSTM layer.
A second approach, ELMo+Dense, exploits a fixed representation of 1024 for the input text: this is obtained by computing a mean-pooling between all contextualized word representations in order to get a trainable feature vector with dimensions $[batch\_size, 1024]$, a dropout layer with rate $0.2$ is followed by two dense layers for prediction. 
For the prediction model ReLU is used as activation function, except for the output layer where we apply sigmoid activation for ELMo+FC and softmax activation for ELMo+LSTM.

\subsubsection{BERT}
We used BERT pre-trained language representations and chose second-to-last hidden layer output as representation for every token of our sentence. We discard the last hidden layer since it contains a bias on the target function where the pre-training was made on. We take all tokens vectors for every sentence and apply average pooling in order to get a fixed representation of the text, obtaining a $768$ dimensional vector. Then, in order to perform mood prediction, we use a very simple classifier that consists in one hidden layer which takes the embedded sentence as input and returns the output of a softmax layer as prediction.

\subsection{Audio-based}

Audio-based classification tasks are recently performed with the use of Convolutional Neural Networks. ConvNets are widely used in the field of computer vision but can be very accurate also for audio classification, since the input to the network can be a 2D representation of the audio such as a Spectrogram or Mel-Spectrogram.
A Spectrogram is a visual representation of the frequency content of a signal that varies over time. For the purpose of this work we will use the Mel-Spectrogram, which is also a time-frequency representation of the audio, but sampled on a Mel-frequency scale, resulting to be a more compact yet very precise representation of the Spectrum of a sound, that has been proven to be more suited when working with CNN for Music Information Retrieval tasks \cite{Choi:2017_2}.

\subsubsection{CNN Architecture}
The network is a $5$ layers Convolutional Neural Network adapted by \cite{Choi:2016}.
The input to the network are $30$ seconds long excerpts of audio, resampled to a unified sampling rate of $12kHz$. The first layer is the Mel-Spectrogram layer that outputs mel-spectrograms in 2D format with $128$ mel bands, $n\_dft$ $512$, $n\_hop$ $256$ \cite{Choi:2017}, followed by a frequency axis batch normalization. After normalization there are $5$ convolutional blocks with 2D convolution, batch normalization, ELU activation and a max pooling operation at the end of each of them.
The output section comprises a flattening of the output followed by a dense layer with a softmax activation function.
The input to the CNN is the output of the Wave-U-Net \cite{Stoller:2018}, a one-dimensional time domain adaptation of the U-Net architecture. It is a convolutional neural network applied on the raw audio waveform used for end-to-end source separation tasks. Here the audio is processed by repeated downsampling blocks and convolution of feature maps.
\section{Experiments and Results}\label{sec:results}

We have conducted all the experimental trainings on the AWS Machine Learning Sagemaker\footnote{\texttt{https://aws.amazon.com/sagemaker/}} infrastructure with the use of GPU multiple instances.

\subsection{Training details}

For all models (text and audio based) we estimated the parameters by minimizing the multi-label cross entropy loss
\begin{equation}
\mathcal{L}(\theta) = \frac{1}{N} \sum_{n}^{N} \Big[-\sum_{i}^{C} y_i^n \log x_i^n\Big]
\end{equation}
where $N$ is the number of examples, $C$ the number of classes, $y^n$ the one-hot vector representing the true label and $x^n$ the model prediction vector using the softmax function.
For the audio-based classification we used the Adam \cite{Kingma:2015} optimizer algorithm with learning rate of $5e^{-3}$, $50$ epochs and early stopping of $15$ epochs. The data batch size was set to $64$. \newline
For all text classification models we use Adam optimizer; training epochs range from 20 to 30 depending on the task, whereas batch-size and learning rate are fixed at $32$ and $1e^{-3}$.

\subsection{MER using Lyrics data}

\begin{table}[ht]
\centering
\begin{tabular}{l|llll}
& \small A & \small P & \small R & \small F1\\ \hline
\small fT+GRU+attn  & 69.88 & 49.70 & 37.06 & 37.62 \\ 
\small fT+Bi-GRU+attn & 77.01 & 52.24 & 46.13 & 47.74 \\ 
\small fT+LSTM+attn & 80.12 & \textbf{69.00} & \textbf{65.91} & \textbf{67.13} \\ 
\small fT+Bi-LSTM+attn  & \textbf{80.81} & 67.98 & 65.24 & 66.41 \\
\small ELMo+2-Dense  & 67.62 & 55.40 & 42.93 & 45.84 \\ 
\small ELMo+LSTM & 74.86 & 59.43 & 47.41 & 50.26 \\ 
\small BERT+Dense & 42.36 & 45.67 & 38.67 & 41.87
\end{tabular}
\caption{Lyrics-based classification results. fastText embedding (fT) is integrated with two types of RNN cells, both as single and bi-directional configuration. For ELMo we apply both a recurrent (ELMo+LSTM) and a fully-connected (ELMo+2-Dense) architecture; lastly for BERT word representations, we use a single layer for prediction. }
\label{table:5}
\end{table}

The results in Table \ref{table:5} show accuracies, average precisions, average recalls and average F1-scores for every approach. It is important to highlight that these metrics are not weighted with the distribution of the classes in Table \ref{table:1}. In this part we can see that fastText embedding in combination with single or bilinear LSTM with attention has appreciable results compared to other approaches. This means that LSTM seems to better learn the emotional context inside a time-synchronized lyrics compared to GRU cells. This is confirmed also analyzing experiments with ELMo embedding, in which two simple layer configurations are applied and LSTM outperforms dense layers by 5 points in accuracy; furthermore, ELMo with LSTM is able to give good results considering the simpler network configuration compared to RNN, as it has no attention module or other additional layers. 
Another focal point is that LSTM+attn with fastText gives approximately the same results of the bidirectional version; therefore, we can conclude that the application of an additional LSTM to analyze text also in reverse order does not provide significant improvements for emotion classification.
For BiLSTM+attention that represents the method achieving best results in terms of accuracy, we provide in Table \ref{table:6} the confusion matrix. 
The results show that the problem of unbalanced dataset reflects some problems in accuracy for classes where there is a small number of examples.

\begin{table}
\centering
\begin{tabular}{c|ccccc}
 & \rotatebox[origin=c]{90}{Sad} & \rotatebox[origin=c]{90}{Joy} & \rotatebox[origin=c]{90}{Fear} & \rotatebox[origin=c]{90}{Anger} & \rotatebox[origin=c]{90}{Disgust}  \\
\hline
\multicolumn{1}{c|}{Sad} &\textbf{82.53} &10.43 &4.27 &2.07 &0.06 \\

\multicolumn{1}{c|}{Joy} &9.84 &\textbf{86.07} &2.35 &1.16 &0.05  \\

\multicolumn{1}{c|}{Fear} &19.79 &10.73 &\textbf{66.88} &1.66 &0.09  \\

\multicolumn{1}{c|}{Anger} &23.48 &12.03 &6.30 &\textbf{54.78} &3.38 \\

\multicolumn{1}{c|}{Disgust} &25.87 &16.93 &9.58 &8.30 &\textbf{39.29} \\
\end{tabular}
\caption{Confusion Matrix as result of fT+BiLSTM+attn method.  }
\label{table:6}
\end{table}

\subsection{MER using Audio data}

\begin{figure}
  \includegraphics[width=\linewidth]{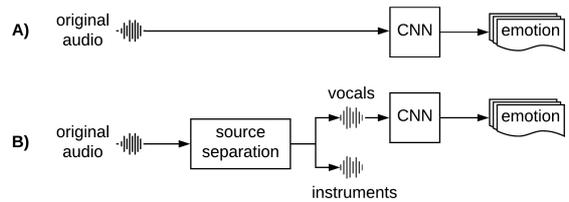}
  \caption{Audio-based classification pipeline. In (A) the original audio (mixture of vocals and instruments) is the input to the CNN. In (B) the original audio is separated into a vocal signal and a mixture of instruments signal by means of a Wave-U-Net, then the voiced signal only is given as input to the CNN.}
  \label{fig:diagram_audio_01}
\end{figure}

We performed two main experiments by maintaining the same network (as described in section \ref{sec:models}) and by only changing the pre-processing methods.
Keeping in mind we only consider and analyze portions of audio related to their corresponding lyric line, as described in section \ref{sec:dataset}, and that we start from a lyrics labeled dataset, in the first experiment we feed the network with mixture of audio and vocals, in the second one, the audio segments are the input of a Wave-U-Net \cite{Stoller:2018}, that is trained at separating vocals from a mixture of instruments. From the output of the Wave-U-Net we retain the audio segments that contain only the vocal track of the song, these are then fed to the CNN. Mel-Spectrogram computation is performed inside the network at its first layer and the Network is trained in an end-to-end fashion starting from the Mel-Spectrogram parameters until its last layer. The pre-processing flow is depicted in Figure \ref{fig:diagram_audio_01}.
One of the difficulties of working with a crowd-based dataset is that examples are highly unbalanced, in fact, the most frequent tag (\textit{sadness}) appears $43.9\%$ of the times, while the least used one (\textit{disgust}) has a percentage of appearance of only $2.1\%$, as shown in Table \ref{table:moods}, therefore we decided to discard the latter, since it was less than one order of magnitude with respect to the other classes. We then used the sample weighting option during training.

\begin{table}[!htbp]
\centering
\begin{tabular}{l|lll}
 & P &  R &  F1 \\ \hline 
Anger    & 33.89 & 10.52 & 5.12 \\ 
Fear     & 9.09 & 8.69 & 8.88  \\ 
Joy      & 39.60 & 31.25 & 34.93  \\ 
Sadness  & 43.22 & 39.23 & 41.12
\end{tabular}
\caption{Audio-based classification results for mixed audio (instruments + vocals) experiment.}
\label{table:audio_results_mixed_01}
\end{table}

\begin{table}
        \centering
        
        \begin{tabular}{c|cccc}
             & \rotatebox[origin=c]{90}{Anger} & \rotatebox[origin=c]{90}{Fear} & \rotatebox[origin=c]{90}{Joy} & \rotatebox[origin=c]{90}{Sadness}  \\
               \hline
               \multicolumn{1}{c|}{Anger} &  \textbf{10.52} & 10.52 & 26.31 &  52.63 \\
           
               \multicolumn{1}{c|}{Fear} &   17.39 &  \textbf{8.69}  & 39.13 &  34.78  \\
            
               \multicolumn{1}{c|}{Joy} &    21.09 & 9.37  & \textbf{31.25} &  38.28  \\
            
               \multicolumn{1}{c|}{Sadness} & 20.00 & 4.61 & 36.15 &  \textbf{39.23}  \\
            
           \end{tabular}
           \caption{Confusion matrix for mixed audio (instruments + vocals)
           experiment.}
            \label{table:confmatrix_mixed_01}
    
       \end{table}

\begin{table}[!htbp]
\centering
\begin{tabular}{l|lll}
 & P &  R &  F1 \\ \hline
Anger    & 14.12 & 20.87 & 17.42 \\ 
Fear     & 12.01 & 17.29 & 15.30  \\ 
Joy      & 41.0 & 35.01 & 38.24  \\ 
Sadness  & 43.22 & 42.23 & 42.12
\end{tabular}
\caption{Audio-based classification results for vocals only experiment.}
\label{table:audio_results_vocals_01}
\end{table}

\begin{table}[!htbp]
        \centering
        
        \begin{tabular}{c|cccc}
             & \rotatebox[origin=c]{90}{Anger} & \rotatebox[origin=c]{90}{Fear} & \rotatebox[origin=c]{90}{Joy} & \rotatebox[origin=c]{90}{Sadness}  \\
               \hline
               \multicolumn{1}{c|}{Anger} &  \textbf{21.05} & 0.0  & 36.84 &  42.10 \\
           
               \multicolumn{1}{c|}{Fear} &   0.0 &  \textbf{17.39}  & 47.82 &  34.78  \\
            
               \multicolumn{1}{c|}{Joy} &    9.37 &  10.15  & \textbf{35.15} &  45.31  \\
            
               \multicolumn{1}{c|}{Sadness} & 10.0 & 11.53 & 36.15 &  \textbf{42.30}  \\
            
           \end{tabular}
           \caption{Confusion matrix for vocals only experiment.}
            \label{table:confmatrix_vocals_01}
    
       \end{table}

In Table \ref{table:audio_results_mixed_01} we present results for precision, recall and f1-score for the mixed audio experiment, while in Table \ref{table:audio_results_vocals_01} we present the same measures related to the vocals only experiment. Despite the total accuracy in the mixed audio scenario is only of $32\%$, as shown in the confusion matrices in Table \ref{table:confmatrix_mixed_01} (mixed), in Table \ref{table:confmatrix_vocals_01} (vocals) the overall accuracy grows to $36\%$. Even though the results of the audio classification are poor, what appears to be relevant is the fact that using vocals tracks only as input to the CNN improves the overall classification performances. This is a clear indicator that more investigations and studies could be done in this direction. 
\section{Conclusion and Future Works}\label{sec:concl_future_works}

In this work we have presented a new dataset, the \textit{Synchronized Lyrics Emotion Dataset}, containing data regarding start and duration of lyrics in the audio and their corresponding emotion labels. We have used this dataset for performing automatic classification of emotions based on the two main sets of information available when dealing with music: lyrics and audio.

We have analyzed the performances of various text embedding methods for lyrics emotion recognition. In the experiments we used contextual feature extraction like ELMo and BERT in combination with simple classifiers in order to compare their results with approaches that exploit non-contextual embedding and more complex classification models. The results have shown that, using the second approach with fastText, performances are better for lyrics emotion classification. Starting from this outcome, a fusion between a contextual embedding and a more complex prediction model might represent a future challenge in this direction.
As Table \ref{table:5} shows, BERT does not provide comparable results to other approaches: BERT represents a very recent work and this first result might be a starting point for future tests and improvements in order to make it efficient also on our lyrics classification task.

From the audio perspective, we have exploited a new WaveNet approach that enables the possibility of separating the audio file into a vocal track and a mixture of instruments track, and we have concluded that, in this case, using vocals only instead of the complete mixed song improves the performances of the emotions classification.

For future works, we could think of further extending the list of tagged emotion labels and repeating the same experiments using Valence-Arousal values, widening the emotion labels possibilities.

\bibliography{ms}

\end{document}